# Face morphing detection in the presence of printing/scanning and heterogeneous image sources

Matteo Ferrara [1*], Annalisa Franco [1], Davide Maltoni [1]

[1] Department of Computer Science and Engineering, University of Bologna, via dell'Università, 50 – Cesena - Italy
[*]matteo.ferrara@unibo.it

**Abstract: Face morphing represents nowadays a big security threat in the context of electronic identity documents as well as an interesting challenge for researchers in the field of face recognition. Despite of the good performance obtained by state-of-the-art approaches on digital images, no satisfactory solutions have been identified so far to deal with cross-database testing and printed-scanned images (typically used in many countries for document issuing). In this work, novel approaches are proposed to train Deep Neural Networks for morphing detection: in particular generation of simulated printed-scanned images together with other data augmentation strategies and pre-training on large face recognition datasets, allowed to reach state-of-the-art accuracy on challenging datasets from heterogeneous image sources.**

## 1. Introduction

The widespread adoption of biometric identification techniques in the context of identity documents poses some concerns for the possibility of fraudulent misuses. Recent studies [1] [2] [3] [4] revealed that ePassports are particularly sensitive to the so called morphing attack, where the face photo printed on paper and provided by the citizen can be altered. Such attack was first described in [2] in the context of face verification at Automated Border Control (ABC) gates where two subjects cooperate to produce a morphed face image (mixing their identities) in order to obtain a regular travel document that could be exploited by both subjects. Of course, in order to succeed in the attack, the morphed face image must be very similar to one of the two subjects (the one applying for the document) to fool the officer during the issuing process, but at the same time must contain enough features of the hidden subject to enable positive verification at the gate for both individuals.

The feasibility of this attack has been analyzed and confirmed by several researchers and some police agencies, thus making the development of proper countermeasures quite urgent.

One of the main challenges for the development of effective solutions for morphing detection is that typically the id photo, natively digital, is printed by the photographer and then scanned by the officer to be stored into the document chip. This Printing/Scanning process (P&S) alters the image information, removing most of the fine details (i.e. digital processing artifacts) that could help to detect morphing. Some preliminary studies, more widely discussed in the next section, show that morphing detection from digital images can be addressed to some extent, but P&S images are still difficult to manage [5]. Promising solutions have been recently obtained by using Deep Neural Networks (DNN), which proved to effectively detect and recognize faces in uncontrolled scenarios [6]. However, to reach a good accuracy, DNN typically require a large training dataset. Unfortunately, in the context of morphing detection, it is difficult to collect large databases of samples: manually producing high quality morphed images is in general a boring and time-consuming activity. Moreover, due to the need of detecting morphing from P&S images, the costs/efforts for printing the images and scanning them again must be also considered. For this reason, most of the approaches in the literature exploit pre-trained deep networks as feature extractors and build on the top of them traditional classifiers (e.g., SVM) that can be trained with relatively small datasets. The aim of this study is to investigate the possibility of artificially generating large sets of morphed images to train DNNs. In particular, this work focuses on the simulation of the P&S process which, coupled with the automatic generation of morphs, can produce large datasets for i) training new networks from scratch or ii) fine-tuning pre-trained DNNs such as AlexNet [7] or VGG [8]. Moreover, an extensive analysis of the network behavior with respect to bona fide/morphed and digital/printed-scanned images enables a deeper understanding of the most relevant image features exploited for classification.

The rest of the paper is organized as follows: Section 2 discusses the state of the art; in Sections 3 and 4, the procedure for automatic printed/scanned image generation and the databases used for train and test are described, respectively. The DNNs used for the experiments are briefly introduced in Section 5 and the experimental results are reported and commented in Section 6. Finally, Section 7 draws some conclusions and discusses possible future research directions.

## 2. Related works and contribution

Although face morphing detection is a recently emerged research area, an increasing number of researchers are working on this topic and the related literature is constantly growing [9]. Existing techniques can be mainly framed in two categories:

- *single-image based*, where the presence of morphing alterations is detected on a single image, such as the id photo presented to the officer at enrolment time or the face image read from an e-document during verification at the gate; *image-pair based* (a.k.a. *differential morphing detection*), where the comparison between a live image (e.g., acquired at the gate) and that stored on the e-document is exploited for morphing detection.

Several literature approaches belong to the first category. The works based on handcrafted features mainly try to analyze the small image degradations produced by the morphing process. In [10] the authors propose a technique for morphing detection based on the analysis of micro-texture variations using Binarized Statistical Image Features (BSIF): an SVM classifier is trained to discriminate bona fide/morphed faces. The authors of [11] argue that the morphed images are characterized by a different texture with respect to the unaltered ones and that a progressive JPG compression can further highlight this aspect; the image content is finally represented by different corner features exploited for classification. In [12] [13] [14] morphing detection is based on Benford features extracted from quantized DCT coefficients, in [15] key-points features (such as SURF, ORB, FAST, etc.) are used, while in [16] [17] [18] [19] texture features such as LBP or Binarized Statistical Image Features (BSIF) are analyzed. An interesting outcome of [16] is that low-level features are not robust when used in cross-database testing or in the presence of simple image manipulations (e.g., rescaling). The authors of [20], [21] and [22] exploit the principle of image source identification for morphing detection, observing that a morphing is a computer-generated image and its sensor-pattern noise is different from that of a real image. Other works make use of topological analysis of facial landmarks to detect alterations introduced by morphing [23] [24]; the idea is interesting in principle, but overall the results obtained are unsatisfactory for real application. Most of the referred approaches, when tested on digital images only, provide good classification performance, but the use of different databases and different evaluation metrics make a comparison quite difficult.

Deep learning techniques based on Convolutional Neural Network (CNN) have been proposed for face morphing detection [17] [25] [26] [27]. The authors of [17] evaluate some networks, pre-trained for face recognition, as feature extractor for digital images, without performing any fine-tuning on the specific morphing detection task, while in [25] two pre-trained networks, AlexNet [7] and VGG19 [8], are used for feature extraction after a fine-tuning step. The authors perform tests on both digital and P&S images and the experimental results clearly confirm that the second type of images represent the main challenge for morphing detection. In [26] some CNNs are used for morphing detection from digital images; the accuracy of pre-trained networks is compared to that of networks learned from scratch, finally leading to the conclusion that pre-trained networks are more robust for this task. The authors of [27] analyze the accuracy of pre-trained networks against semantic (partial morphing on some specific face regions) and black box attacks (partial occlusions), and highlight, for the two kind of images, the most relevant regions analyzed by the networks for classification. Finally the authors of [28] combine features of different nature, hand-crafted and extracted by CNNs, demonstrating that a substantial improvement in detection performance can be achieved by their integration.

A limited number of approaches perform morphing detection by image-pair comparison. The first approach has been introduced in [29] [30] where the inverse process of morphing (called demorphing) is adopted to revert the effects produced by morphing. The demorphing technique proved to be effective both on digital and P&S images. The same detection scheme has been considered in [31] where different features are evaluated both for single-image and differential morphing detection. Deep features extracted from different networks are used in [32] for image comparison, while the authors of [33] exploit features from 3D shape and the diffuse reflectance component estimated from the image. Finally, a landmark-based morphing detection approach is proposed in [34] to compare bona fide and suspected morphed images.

Overall, an analysis of the literature allows to identify two major challenges for morphing detection techniques: i) robustness to the P&S process; ii) ability to generalize across different databases [35]. The present work mainly focuses on these two aspects. In particular, this paper provides the following contributions:

- Adoption of a simple P&S simulation model [36] for data augmentation, enabling the possibility of producing training images without the cost/effort of the real P&S process. The state-of-the-art about simulating the printing/scanning process in the context of face recognition is very limited. To the best of our knowledge the most relevant paper [37] has been proposed very recently. The approach exploits generative networks to simulate the real process, providing interesting results from the visual point of view. This techniques requires a training stage based on real P&S images; on the contrary, an advantage of the model used in our work is that no real P&S images are needed for training and a variety of devices or acquisition conditions can easily be simulated just varying the main algorithm parameters. The experimental results will show that such simulation produces a significant performance improvement on morphing detection from P&S images.
- Extensive experiments using four different well-known DNN architectures on several test datasets and public benchmarks.
- Thorough performance evaluation on several public benchmarks and comparison with state-of-the-art techniques.
- Experimental results confirming the feasibility of the print/scan simulation model proposed here to deal with real P&S images.

## 3. Automatic image generation

In order to exploit the great potential of CNNs for classification, a very large set of images is typically needed and usually data augmentation techniques are applied [38] to increase the number of samples available for training; geometric and photometric transformations are the most frequently adopted modifications. In the context of morphing detection, the network training requires both real and morphed image samples, possibly in the two formats (digital and P&S). To avoid the effort/cost of collecting a large dataset we proposed novel techniques for automatically generating high quality morphed face images (Section 3.1) and simulating the P&S process (Section 3.2).

### 3.1. Face morphing

Morphed images can be obtained quite easily using one of the many existing tools and plugins (e.g., Sqirlz Morph [39]). However, the systematic generation of morphed images with specific characteristics can be better realized by ad hoc techniques. Here we adopt the approach described in [29] which includes an automatic image retouching phase to minimize visible artifacts. Given two images $I_0$ and $I_1$, the process generates a set of frames $\mathbb{M} = \{I_\alpha, \alpha \in \mathbb{R}, 0 < \alpha < 1\}$ representing the transformation of the first image ($I_0$) into the second one ($I_1$) (see Figure 1). In general, each frame is a weighted linear combination of $I_0$ and $I_1$, obtained by geometric warping of the two images based on corresponding landmarks and pixel-by-pixel blending. Formally:

$$I_\alpha(\mathbf{p}) = (1-\alpha) \cdot I_0\left(w_{P_\alpha \to P_0}(\mathbf{p})\right) + \alpha \cdot I_1\left(w_{P_\alpha \to P_1}(\mathbf{p})\right), \quad (1)$$

where:

- $\mathbf{p}$ is a generic pixel position;
- $\alpha$ is the frame weight factor (representing the presence of the two contributing subjects);
- $P_0$ and $P_1$ are the two sets of landmarks in $I_0$ and $I_1$, respectively;
- $P_\alpha$ is the set of landmarks aligned according to the frame weight factor $\alpha$;
- $w_{B \to A}(\mathbf{p})$ is a warping function.

A number of different morphed images can be obtained according to the value of the weighting factor $\alpha$ (i.e. the weight of the two subjects in the combination) as shown in Figure 1.

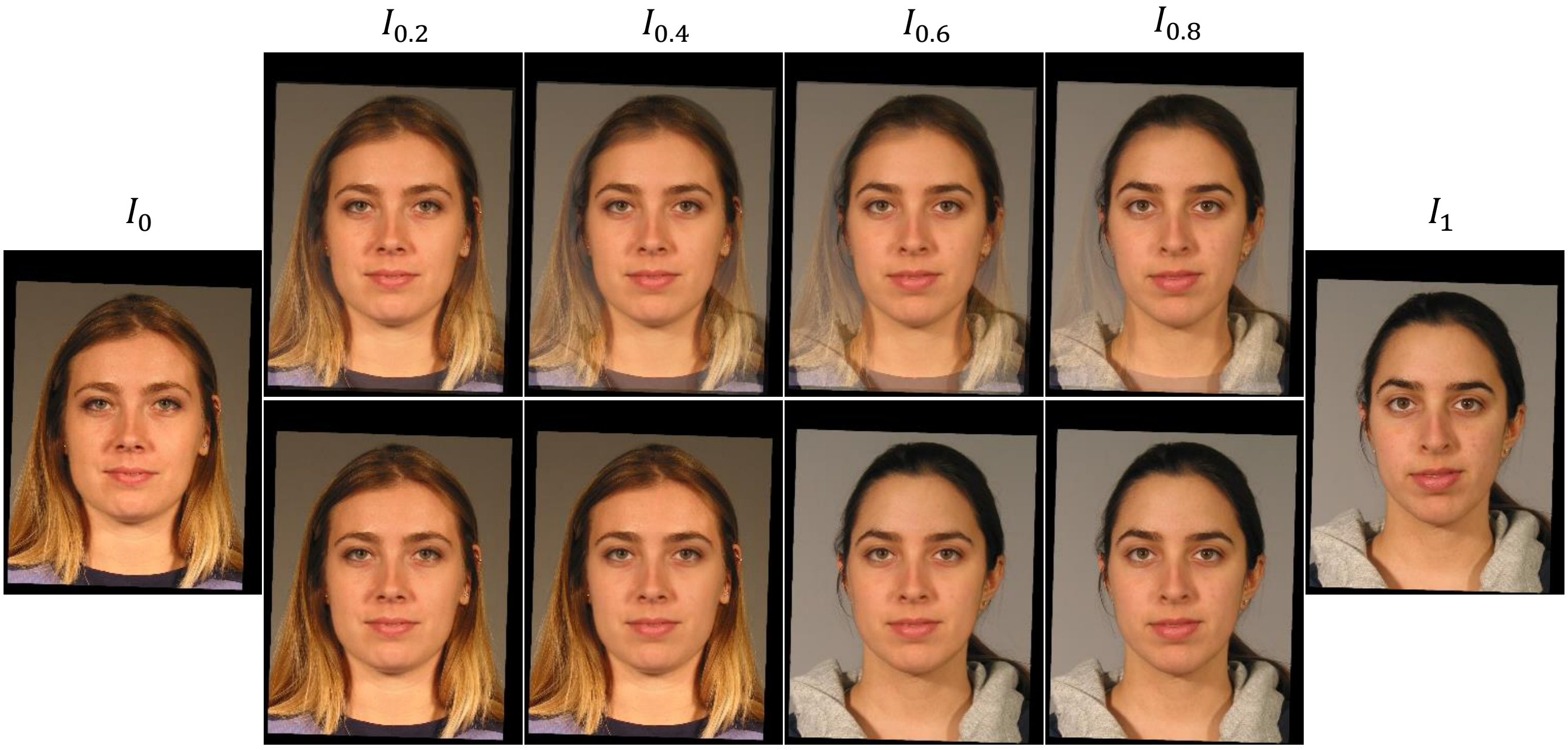

***Figure 1.*** *Example of morphed frames obtained by the morphing procedure, gradually moving from $I_0$ to $I_1$ (first row). In the second row the result of the automatic retouching process used to remove visible artifacts is shown.*

### 3.2. Modelling the printing and scanning process

The P&S process is quite complex: i) digital images are first conveyed to the physical, continuous domain and then ii) re-digitalized and discretized by the scanning process. The image alterations introduced involve both pixel value distortions (i.e. luminance, contrast and gamma corrections, chrominance variations and blurring of adjacent pixels) as well as minor geometric alterations due to the positioning on the scanner surface.

Focusing on the pixel value distortion, according to the model proposed in [36] the P&S process of a generic digital image $I$ produces a modified, discrete version of the image $\tilde{I}$ as:

$$\tilde{I}(\mathbf{p}) = K\left[I(\mathbf{p}) * \tau_1(\mathbf{p}) + \left(x(\mathbf{p}) * \tau_2(\mathbf{p})\right) \cdot N_1\right] \cdot s(\mathbf{p}), \quad (2)$$

where:

- Function $K$ represents the responsivity of the acquisition device;
- $s(\mathbf{p})$ is the sampling function which characterizes the digitalization process of the continuous printed image;
- $\tau_1$ models the system point spread function $\tau_1(\mathbf{p}) = \tau_P(\mathbf{p}) * \tau_S(\mathbf{p})$ where $\tau_P(\mathbf{p})$ and $\tau_S(\mathbf{p})$ represent the point spread function of printer and scanner, respectively;
- $\tau_2$ is a high-pass filter used to represent higher noise variance near the edges;
- $N_1$ is a white Gaussian random noise.

The following responsivity function $K$ is adopted:

$$K(x) = \omega \cdot (x - \beta_x)^\gamma + \beta_K + N_2(x) \quad (3)$$

It includes color adjustments coefficients ($\beta_x$ and $\beta_K$), gamma correction ($\gamma$) and a noise component $N_2(x)$ whose power is related to pixel intensity (usually higher noise on dark pixels is observed due to the different sensors' sensitivity to the image reflectivity).

Due to some device-dependent unknown parameters, the adaption of this model to real cases is not straightforward. In particular, the point spread functions of the devices ($\tau_P$ and $\tau_S$ in Eq. (2)) are not available, and they are approximated by two Gaussian blurring filters of size $k_1$, $k_2$ and standard deviation $\sigma_1$, $\sigma_2$.

The model is quite flexible and allows to modify different image characteristics, related to both visual quality and low-level signal content. Figure 2-Figure 5 show the impact of the different model parameters on the result. In particular, $\omega$ mainly controls the image contrast and brightness (see Figure 2), while the overall system gamma, i.e. the combined effect of all gamma values applied to the imaged by the printing/scanning devices, can be adjusted by properly tuning $\gamma$ (see Figure 3). Further variations to image color and saturation can be obtained though $\beta_K$ and $\beta_X$ parameters (see Figure 4). Finally the parameters of the Gaussian smoothing filter ($k$ and $\sigma$) produce the most evident modification introduced by the P&S process, i.e. the blurring effect represented in Figure 5.

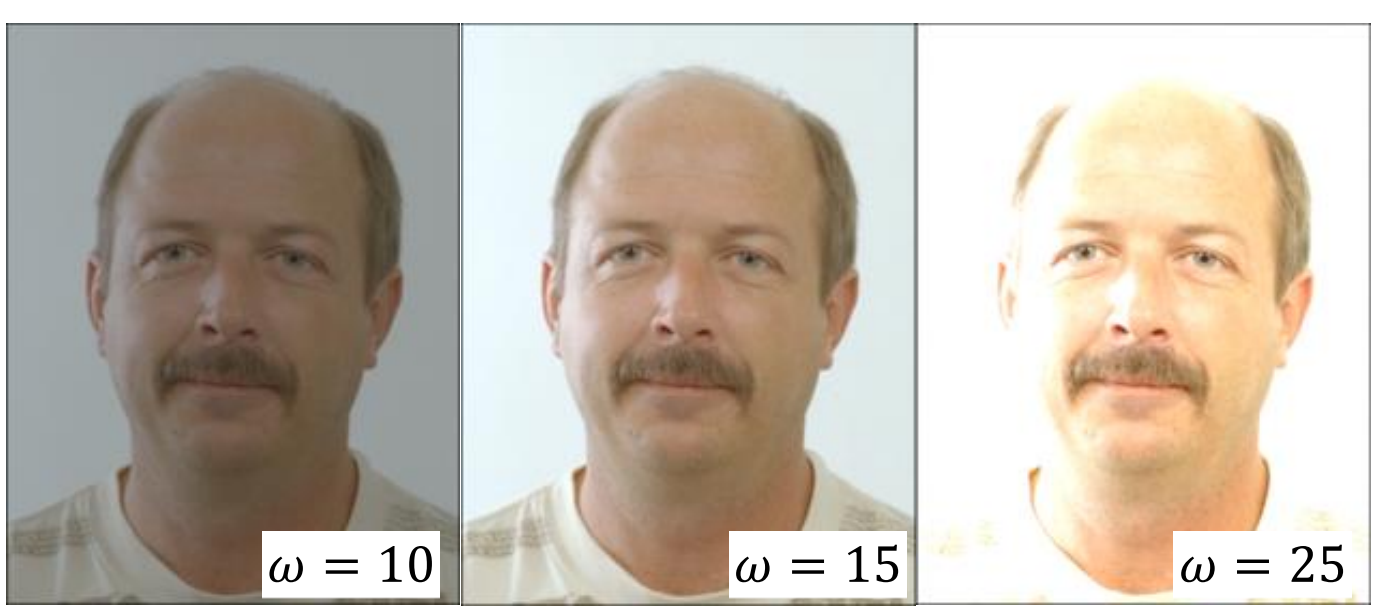

***Figure 2.*** *Variation of ω parameter in the P&S simulation model applied to Figure 6.(a): this parameter mainly affects image contrast and brightness.*

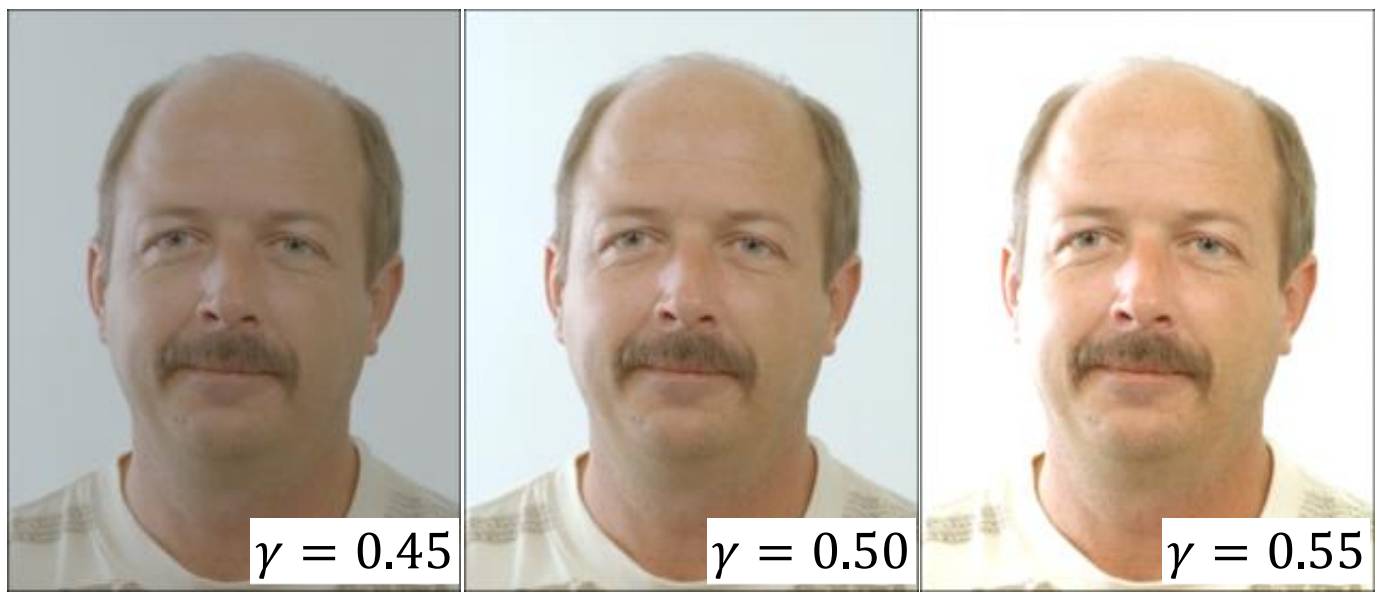

***Figure 3.*** *Variation of* $\boldsymbol{\gamma}$ *in the P&S simulation model applied to Figure 6.(a): this parameter regulates the gamma corrections produced by the printing and scanning devices.*

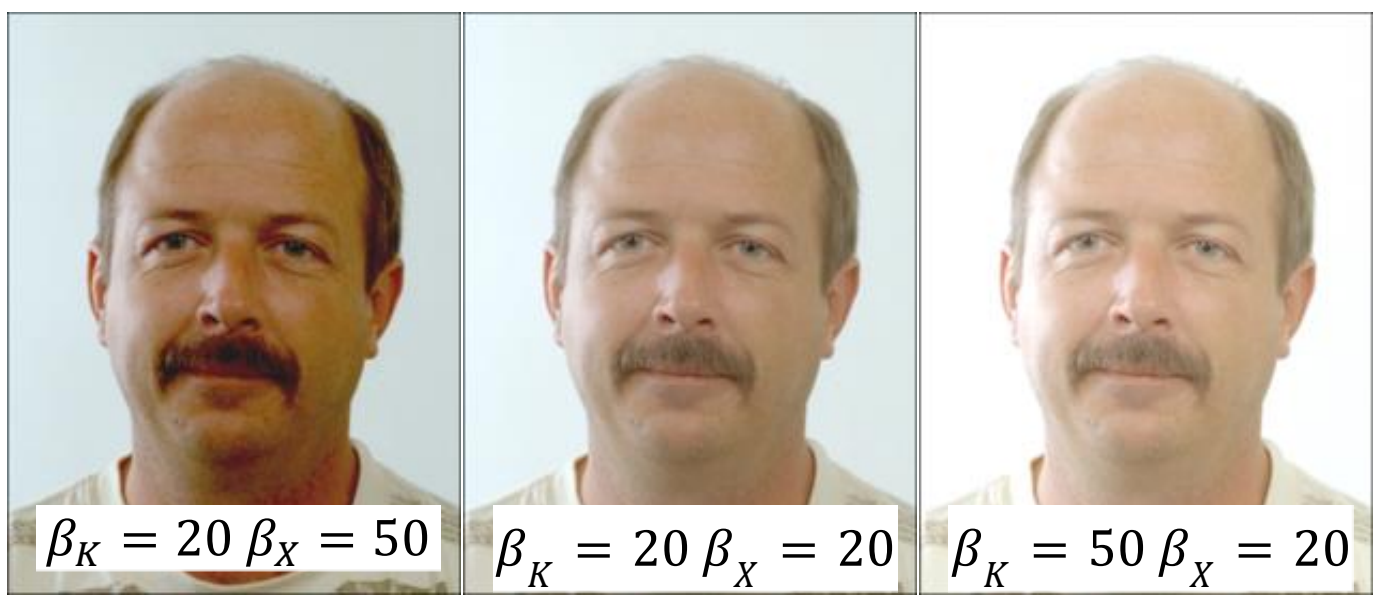

***Figure 4.*** *Variation of* $\boldsymbol{\beta_K}$ *and* $\boldsymbol{\beta_X}$ *in the P&S simulation model applied to Figure 6.(a): these parameters control the image color and saturation.*

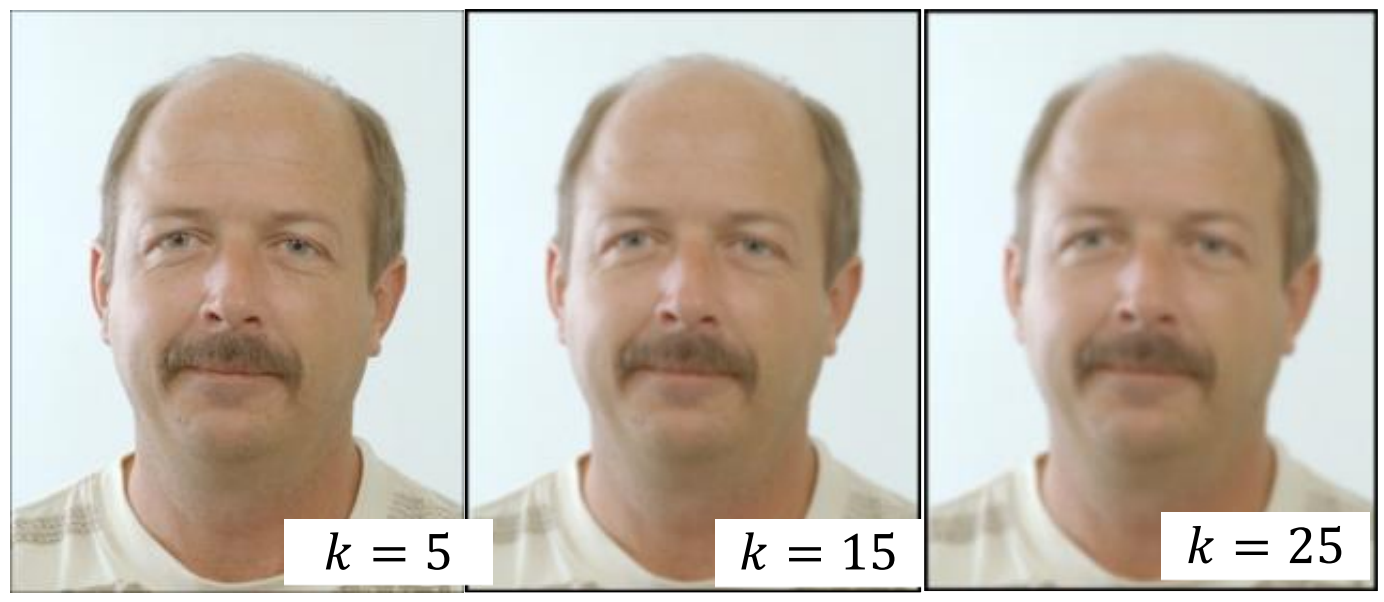

***Figure 5.*** *Variation of* $\boldsymbol{k}$ *in the P&S simulation model applied to Figure 6.(a): this parameter controls the amount of image blurring.*

In Figure 6 a real P&S image is compared with a simulated P&S image of the same digital image. The image spectrum is also reported to appreciate the low-level signal modifications produced by the P&S process. As clearly visible in the example, the digital image is much richer of fine details (high frequencies) which are noticeably attenuated after P&S. The spectrum of the simulated P&S image (Figure 6.(f)) is quite similar to that of the real one (Figure 6.(e)). We can quantify the similarity between the image spectra adopting commonly used metrics such as the spectral angle [40] (a measure of distance between two spectra) or the correlation value. If we compare the digital image and the real P&S of Figure 6, the spectral angle is quite high (0.69) with a correlation value of 0.77. The similarity between the real P&S and the simulated one is much higher, as confirmed by the smaller spectral angle (0.38) and a higher correlation value (0.93).

The parameters used for image generation (see Table 1) have been chosen in order to produce images visually similar to the real P&S ones (*MorphDB*$_{P\&S}$ database described in Section 4.2), but no specific optimizations have been carried out (see Figure 6).

***Table 1.*** *Parameter values used in for P&S simulation.*

| **Parameter** | $\omega$ | $\beta_X$ | $\beta_K$ | $\gamma$ | $k_1, k_2$ | $\sigma_1, \sigma_2$ |
|---|---|---|---|---|---|---|
| **Value** | 15.5 | 20 | 20 | 0.5 | 3 | 1.2 |

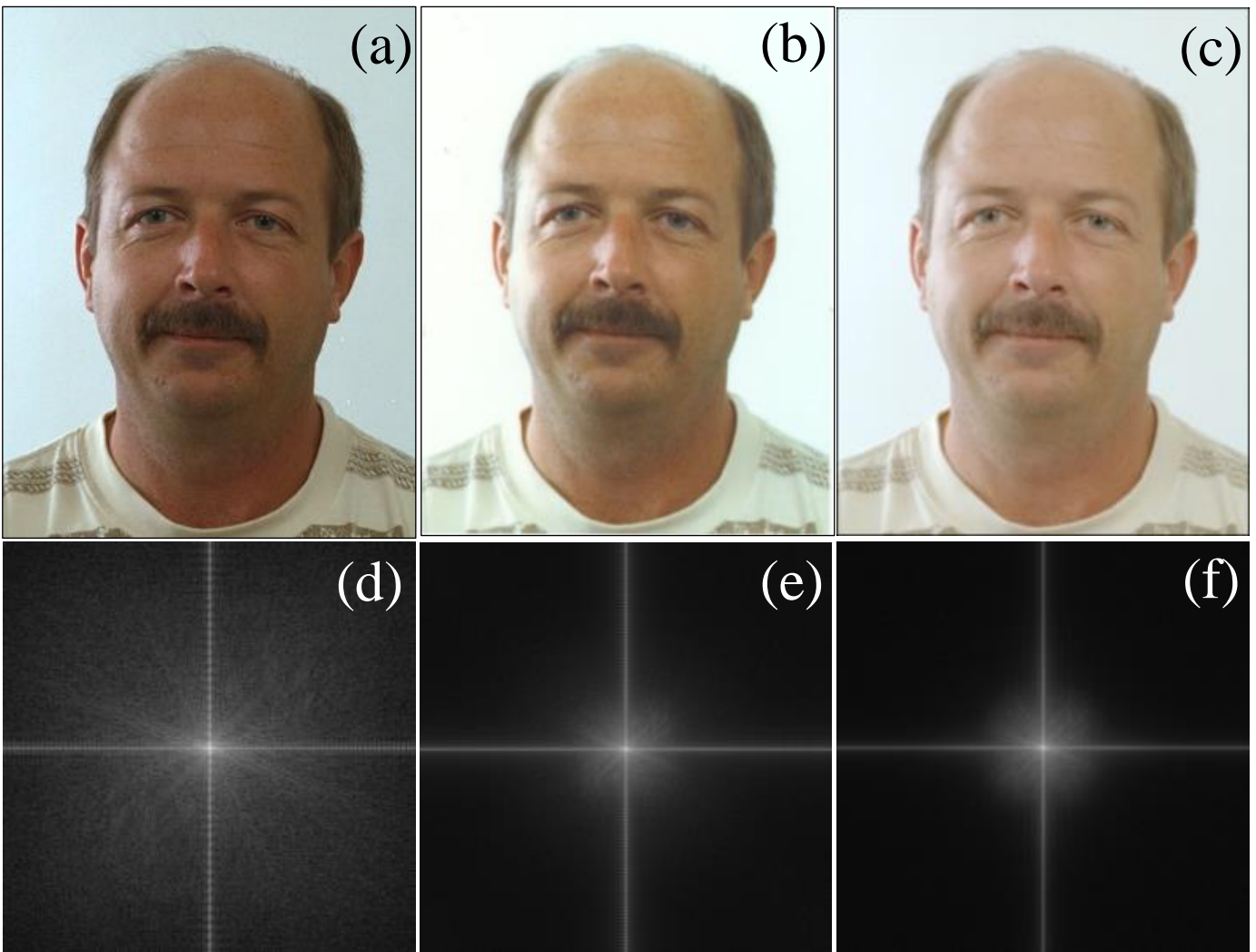

***Figure 6.*** *For the digital image (a) the result of the real (b) and simulated (c) P&S processes is provided. The corresponding image spectrum is also given for the digital image (d), the real (e) and the simulated P&S (f).*

## 4. Databases

### *4.1. Training sets*

For network training we used the Progressive Morphing Database (PMDB) described in [29]. It contains 6000 morphed images automatically generated starting from 280 different subjects selected from the AR [41], FRGC [42] and Color Feret [43] [44] databases using different morphing factors ($\alpha \in \{0.1, 0.15, 0.2, 0.25, 0.3, 0.35, 0.4, 0.45\}$ in Eq. (1)).

Since PMDB contains a different number of bona fide and morphed images, a new balanced database (called $Digital$) has been derived as follows:

1. two images of each subject are chosen resulting in 560 bona fide images;

2. 560 morphed images are randomly selected from the PMDB morphed images.

The P&S process has been simulated by applying the procedure described in Section 3.2 on all $Digital$ images; we will refer to this dataset as $\widetilde{P\&S}$.

### 4.2. Test sets

The models trained on the datasets introduced in Section 4.1 are then tested on the following databases:

- $MorphDB_D$ [29]: it consists of 130 bona fide images (not morphed) and 100 morphed images (50 males and 50 females) produced with a significant manual intervention in order to minimize visible artifacts (see Figure 7).
- $MorphDB_{P\&S}$ [29]: P&S version of $MorphDB_D$. The images have been printed on high quality photographic paper by a professional photographer and then scanned (see Figure 8).
- NIST FRVT-MORPH benchmark [45] including several image subsets of variable quality. The morphed images in this dataset have been generated with a plurality of morphing algorithms, thus representing a very hard challenge.
- SOTAMD benchmark [46], including high quality morphed images (both digital and printed/scanned).

The $MorphDB_D$, $MorphDB_{P\&S}$ and SOTAMD datasets are publicly available for testing in the FVC-onGoing platform [47] [48], a web-based automated evaluation system for biometric recognition algorithms.

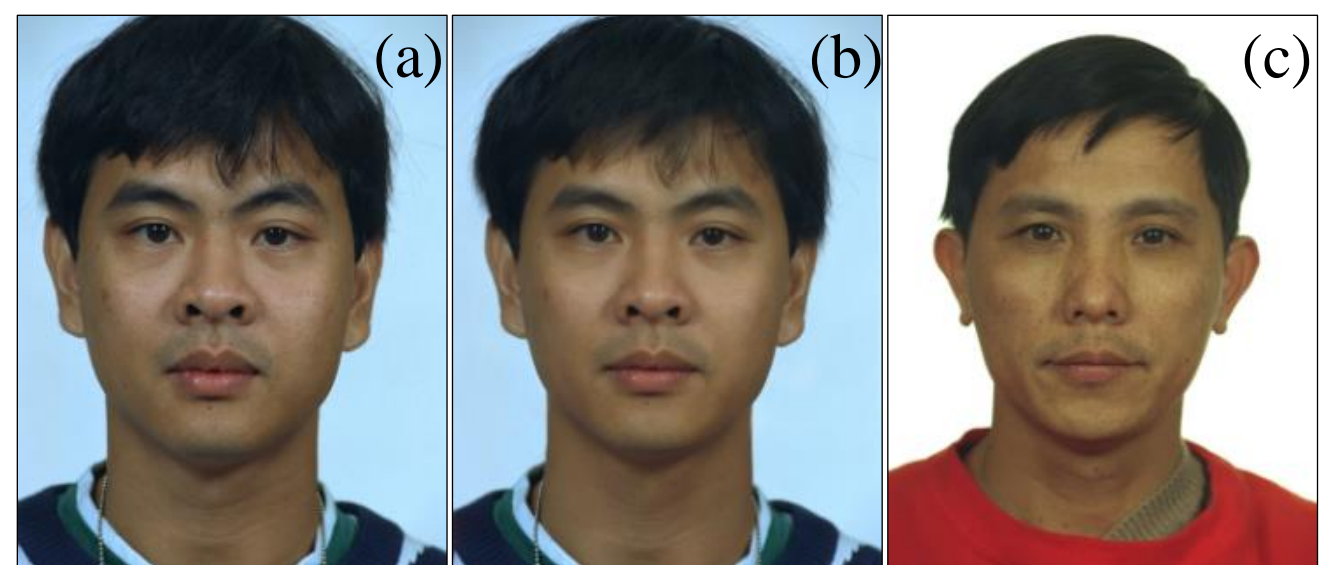

***Figure 7.*** *Images from MorphDB$_D$ database: digital version of bona fide images of two subjects (a) and (c) and the resulting morphed image (b).*

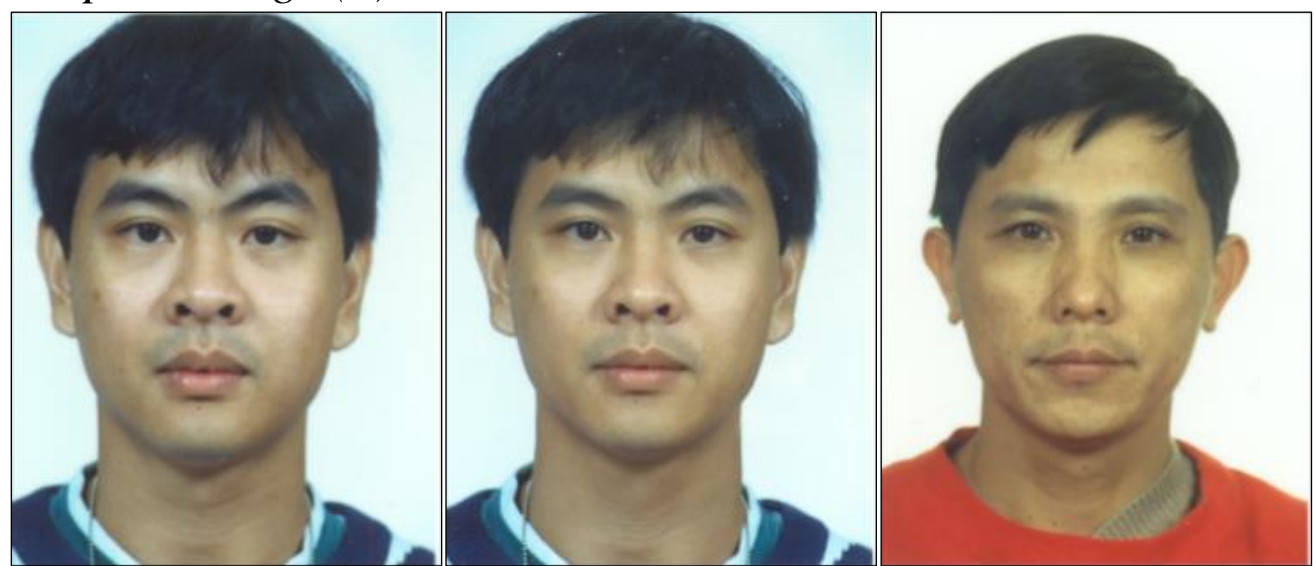

***Figure 8.*** *Images from MorphDB$_{P\&S}$ database: P&S version of the images reported in Figure 7.*

### 4.3. Data normalization

Since face images come from various sources presenting different size and resolution, it is important to normalize them before processing (see Figure 9). For this reason, each image is normalized as follows:

1. the eye centers and the nose tip are detected using Neurotechnology VeriLook SDK 10.0 [49];
2. the image is resized to obtain an eye center distance of 150 pixels;
3. a sub-image of size 350×400 pixels is cropped centered on the nose tip.

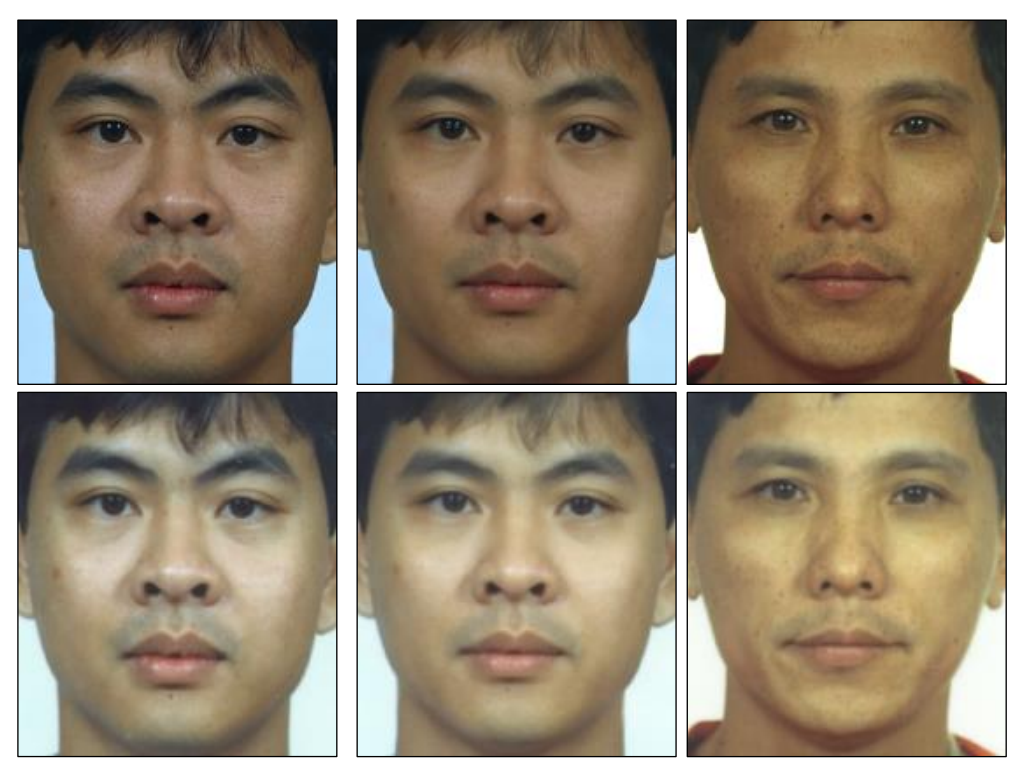

***Figure 9.*** *Normalized images from Figure 7 (first row) and Figure 8 (second row).*

### 4.4. Data augmentation

Both training databases ($Digital$ and $\widetilde{P\&S}$) contain 1120 images, not many for an effective network training. To increase the number of samples, data augmentation is applied obtaining different augmented databases (see Table 2).

In particular, the following transformations are applied:

- horizontal mirroring;
- rotation centered on the nose tip ({-5°,0°,+5°});
- horizontal and vertical translation ({-1,0,+1});
- multi-crop, i.e. extracting from each image (size 350×400) five sub-images corresponding to the four corners and the central region [7]. The crop size is fixed according to the image input size of the specific network (see Section 5). In the tests where multi-crop is not enabled, only the central region is used.

## 5. Deep Neural Networks for morphing detection

In this work we considered different well-known pre-trained deep neural networks (see Table 3). The first two networks, already used for morphing detection in previous works [25] [26], have been trained on natural images (i.e. ImageNet [50]) and therefore the learned filters are not specific for face representation. The last two networks are state-of-the-art models trained on very large face datasets: we can expect that the filters in the low and intermediate levels of these networks are capable of extracting very powerful face feature that can be exploited for morphing detection.

The last layer of all the considered architectures has been changed to deal with a two class problem (morphed vs bona fide): as a consequence, the corresponding weights need to be learned from scratch.

### 5.1. Fine-tuning

Starting from the pre-trained networks, a first fine-tuning step has been performed on $Digital_{Au}$ and $Digital_{Mc}$ datasets, separately, for 5 epochs each. Therefore, for each network architecture, we obtained two differently tuned networks able to detect digital morphed images but presenting poor results on P&S ones (see Section 6.2). To overcome this limit, a second fine-tuning step has been performed on $\widetilde{P\&S}_{Au}$ and $\widetilde{P\&S}_{Mc}$ datasets, for a single epoch each. For both fine-tuning stages, we used SGD optimization with a fixed learning rate of 0.0001 and a momentum of 0.9.

At test time, if multi-crop augmentation were used during training, the prediction probabilities are calculated as the average probabilities across five sub-images (i.e. the four corners and the central region) cropped from the normalized 350×400 image. Otherwise only the central region is used for classification.

***Table 2.*** *Characteristics of the different training datasets.*

| Name | P&S Simulation | Data Augmentation | | | | # Images | | |
|---|---|---|---|---|---|---|---|---|
| | | Horizontal Mirroring | Rotation | Horizontal and Vertical Translation | Multi-crop | Bona fide | Morphed | Total |
| $Digital$ | | | | | | 560 | 560 | 1120 |
| $\widetilde{P\&S}$ | √ | | | | | | | |
| $Digital_{Au}$ | | √ | √ | √ | | 30240 | 30240 | 60480 |
| $\widetilde{P\&S}_{Au}$ | √ | √ | √ | √ | | | | |
| $Digital_{Mc}$ | | √ | √ | | √ | 16800 | 16800 | 33600 |
| $\widetilde{P\&S}_{Mc}$ | √ | √ | √ | | √ | | | |

***Table 3.*** *Neural networks used in the experimentation.*

| Name | Architecture | Pre-trained on | | | Input image size |
|---|---|---|---|---|---|
| | | Image Type | Database name | Database size | |
| AlexNet [7] [51] | AlexNet - BVLC version | Natural | ImageNet [50], specific ILSVRC subsets [52] | 1.2M | $227 \times 227$ |
| VGG19 [8] | VGG – 19 weight layers | | | | $224 \times 224$ |
| VGG-Face16 [53] | VGG – 16 weight layers [8] | Face | VGG-Face dataset [53] | 2.6M | $224 \times 224$ |
| VGG-Face2 [54] | ResNet-50 [55] | | VGG-Face 2 dataset [54] | More than 3M | |

## 6. Experiments

Several experiments have been carried out to evaluate the robustness of DNNs for morphing detection with respect to: i) cross-database testing and ii) P&S images.

### 6.1. Testing protocol and performance indicators

For each experiment bona fide and morphed face images are used to compute Bona fide (BPCER) and Attack Presentation Classification Error Rates (APCER), as defined in [56].

The following performance indicators are calculated:

- Accuracy: the percentage of face images correctly classified as bona fide or morphed;
- Equal-Error Rate (EER): the error rate for which both BPCER and APCER are identical;
- BPCER@APCER= $p\%$ : the lowest BPCER for APCER≤ $p\%$;

Detection Error Tradeoff (DET) curve: the plot of APCER against BPCER.

### 6.2. Results on $MorphDB_D$ and $MorphDB_{P\&S}$

Table 4 reports the results obtained in terms of accuracy, EER, and BPCER (at different levels of APCER) as a function of i) the testing database, ii) the network and iii) the training set used. The corresponding DET curves are available in [57].

The results show a variable behavior over different test databases. The performance measured over the *MorphDB*$_D$ dataset are good for all the evaluated networks, even if here the ImageNet pre-trained models (AlexNet and VGG-19) often achieve the best results. We argue that to detect artifacts and traces of digital manipulations that characterize digital morphed images, the general filters learned from natural images can be even more powerful than specific filters optimized for invariant face recognition. This observation is aligned with the outcomes of [26].

The test on *MorphDB*$_{P\&S}$, allows to evaluate the performance when the P&S process comes into play. In general, the results show that networks trained only on digital images are not able to deal with P&S images; all the architectures suffer from this issue and provide quite bad results. Exploiting simulated P&S images for network training allows in some cases to obtain a significant improvement (e.g., the accuracy of VGG-Face16 network trained with multi-crops grows from about 56% to 93%); these results are quite encouraging if we consider that no real P&S images have been used during training. Overall an accuracy of 85-90% can be reached with reasonable values of EER and BPCER at APCER=10% and 5%. In general, among the different data augmentation techniques, the multi-crop approach provides better results. Looking at the performance of the different networks, here we observe an opposite behavior with respect to the experiments on digital images: in fact, the best performing nets are the VGG-Face models pre-trained on large face datasets with AlexNet and VGG19 struggling to reach decent performance. Since P&S removes most of the digital artifacts we argue that more powerful and problem specific feature detectors are needed to solve such a complex problem.

***Table 4.*** *Performance indicators of the evaluated networks on the testing databases using different training sets. The best result on each test database is highlighted in bold.*

| Test | Net | Training | Accuracy (%) | EER (%) | BPCER (%) at | | |
|---|---|---|---|---|---|---|---|
| | | | | | APCER=10% | APCER=5% | APCER=1% |
| *MorphDB$_D$* | AlexNet | $Digital_{Au}$ | **98.3** | 1.8 | 0.8 | 0.8 | 3.8 |
| | | $Digital_{Mc}$ | 96.1 | 1.3 | 0.8 | 1.5 | 1.5 |
| | VGG19 | $Digital_{Au}$ | 92.2 | 3.9 | 0.8 | 3.8 | 10.8 |
| | | $Digital_{Mc}$ | 94.3 | 4.3 | 0.8 | 3.1 | 5.4 |
| | VGG-Face16 | $Digital_{Au}$ | 93.9 | 3.9 | 0.8 | 1.5 | 10.0 |
| | | $Digital_{Mc}$ | 97.4 | **0.9** | **0.0** | **0.0** | **0.0** |
| | VGG-Face2 | $Digital_{Au}$ | 95.2 | 1.8 | **0.0** | 1.5 | 3.1 |
| | | $Digital_{Mc}$ | 93.0 | **0.9** | **0.0** | 0.8 | 0.8 |
| *MorphDB$_{P\&S}$* | AlexNet | $Digital_{Au}$ | 43.5 | 28.7 | 50.8 | 53.8 | 66.2 |
| | | $Digital_{Mc}$ | 43.5 | 32.7 | 64.6 | 74.6 | 83.1 |
| | | $Digital_{Au} + \widetilde{P\&S}_{Au}$ | 67.4 | 20.9 | 43.1 | 52.3 | 70.0 |
| | | $Digital_{Mc} + \widetilde{P\&S}_{Mc}$ | 83.5 | 13.9 | 25.4 | 41.5 | 77.7 |
| | VGG19 | $Digital_{Au}$ | 47.0 | 32.7 | 57.7 | 71.5 | 89.2 |
| | | $Digital_{Mc}$ | 44.3 | 30.4 | 52.3 | 66.9 | 84.6 |
| | | $Digital_{Au} + \widetilde{P\&S}_{Au}$ | 60.4 | 18.2 | 36.9 | 45.4 | 70.0 |
| | | $Digital_{Mc} + \widetilde{P\&S}_{Mc}$ | 56.5 | 24.8 | 49.2 | 54.6 | 55.4 |
| | VGG-Face16 | $Digital_{Au}$ | 60.4 | 12.7 | 13.8 | 20.8 | 69.2 |
| | | $Digital_{Mc}$ | 56.5 | 11.3 | 12.3 | 22.3 | 63.1 |
| | | $Digital_{Au} + \widetilde{P\&S}_{Au}$ | 89.6 | 7.3 | 7.7 | 15.4 | 39.2 |
| | | $Digital_{Mc} + \widetilde{P\&S}_{Mc}$ | **93.5** | **6.1** | **2.3** | **6.9** | 43.8 |
| | VGG-Face2 | $Digital_{Au}$ | 51.7 | 16.5 | 20.0 | 23.8 | 40.0 |
| | | $Digital_{Mc}$ | 45.7 | 15.7 | 18.5 | 33.1 | 80.0 |
| | | $Digital_{Au} + \widetilde{P\&S}_{Au}$ | 74.3 | 8.2 | 6.2 | 9.2 | 25.4 |
| | | $Digital_{Mc} + \widetilde{P\&S}_{Mc}$ | 86.5 | **6.1** | 4.6 | 7.7 | **17.7** |

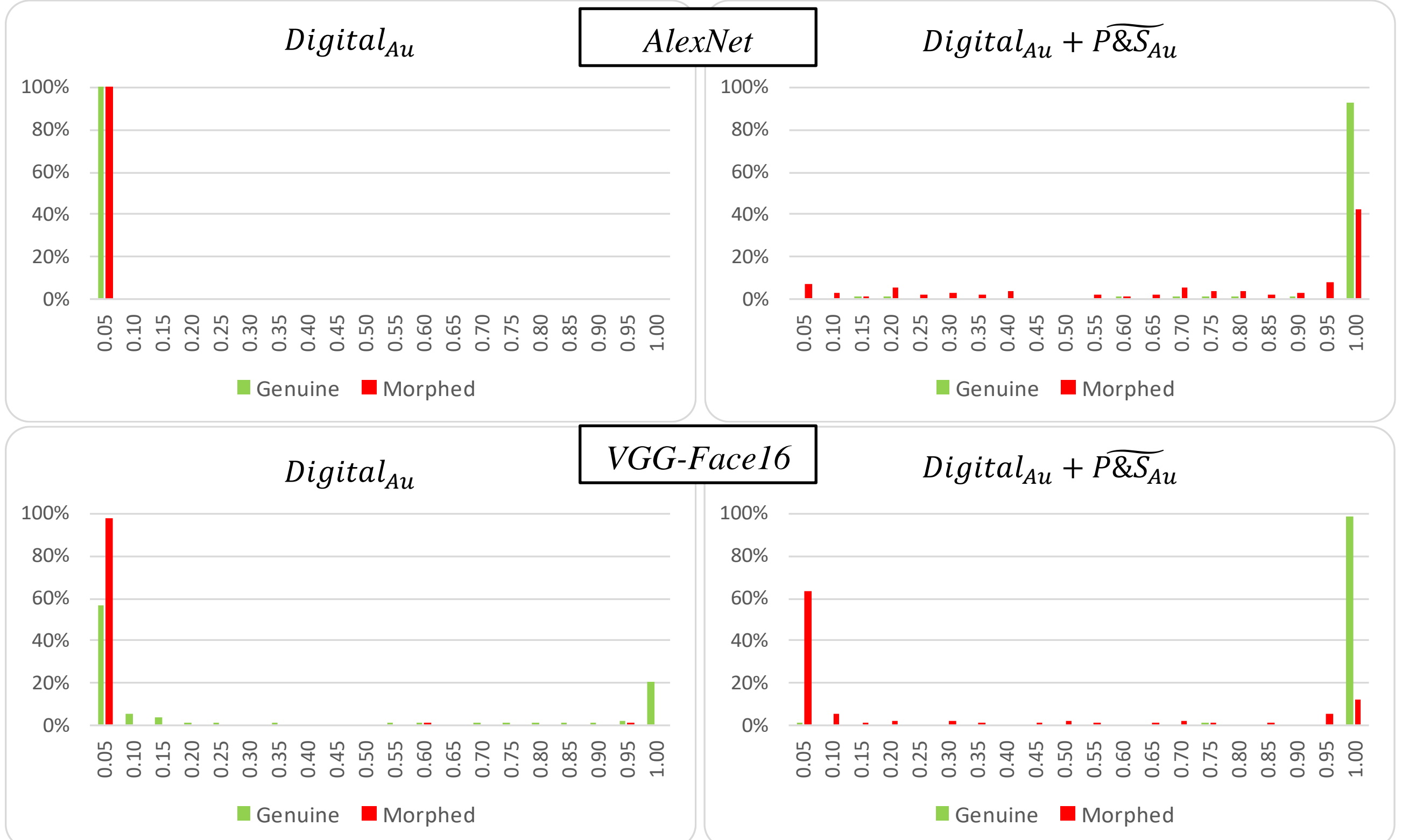

***Figure 10.*** *Bona fide and morphed score distribution on MorphDB$_{P\&S}$ for AlexNet and VGG-Face16 networks obtained using the $Digital_{Au}$ training set (1st column) and the $Digital_{Au} + \widetilde{P\&S}_{Au}$ training set (2nd column).*

To better analyze the effects of extending the digital training set with simulated P&S images the bona fide and morphed score distributions of AlexNet and VGG-Face16 networks trained with the $Digital_{Au}$ and the $Digital_{Au} + \widetilde{P\&S_{Au}}$ training sets are reported in Figure 10. The graphs clearly show that the networks trained on digital images only ($Digital_{Au}$) return a score close to 0 for both bona fide and morphed images. This means that the modifications introduced by P&S remove the textural details that makes bona fide and morphed images distinguishable. When training is extended with simulated P&S images ( $Digital_{Au} + \widetilde{P\&S_{Au}}$ ), the network pre-trained on face images (VGG-Face16) is able to learn P&S specific features making it able to discriminate bona fide from morphed images. Therefore, the bona fide scores become higher, while the morphed scores are generally kept quite low, as clearly visible by the score distributions for VGG-Face16. On the contrary AlexNet does not benefit of this further training step whose introduction determines an increment of all the scores (bona fide and morphed).

### 6.3. Results on NIST FRVT MORPH

Two of the most promising solutions identified in our internal tests (AlexNet trained on $Digital_{Au}$ for the digital images and VGG-Face16 trained on $Digital_{Au} + \widetilde{P\&S_{Au}}$ for the P&S images) have been submitted for evaluation at NIST FRVT MORPH which provides a huge and thorough comparative evaluation of face morphing detection algorithms. Please refer to the report [45] and the evaluation website [58] for the full set of results; in this paper, for lack of space,we report in Figure 11 a subset of the NIST DET plots comparing single-image based detection algorithms on several image subsets (5 digital and 1 printed and scanned). Overall the results show that morphing detection from single images is a very hard task, in particular when heterogeneous datasets are considered. The proposed approach compares favourably with most of the evaluated approaches, and presents overall comparable performance with the *ntnussl_002* algorithm. In the *Lincoln* subset (Figure 11.e) the proposed approach is outperformed by other techniques, even if the best reference value (APCER@BPCER = 0.01%) is reached by the proposed algorithm. In the *Print and Scan* dataset the proposed approach ranks second among the tested algorithms and this is very encouraging if we consider that no real printed/scanned images have been used to train our system; this confirms the efficacy of the simulation procedure proposed here.

The result in Figure 11.d are worth of attention; in this case the morphed images were generated using the morphing algorithm described in section 3.1 so the level of performance achieved is of course very good, significantly better than all the other results. This behaviour confirms the importance of training the system with representative data and suggests that a higher robustness can be achieved extending the training data to a variety of morphing algorithms. This would probably also allow to improve the results on the subsets of Figure 11.e and Figure 11.f.

### 6.4. Results on SOTAMD benchmark

The same solutions tested at NIST have also been tested on the SOTAMD benchmark, which revealed to be a very hard challenge, as confirmed by the very modest results reached by the tested algorithms (see Table 5). The proposed approach achieves an EER value better than the other approaches in the P&S testing set and this is a positive indicator since the SOTAMD P&S images have been produced by multiple processing pipelines reproducing the real workflow used by different countries for passport issuing. However, in general, the BPCER values measured in this benchmark are very bad, confirming that that morphing detection from single images is still an open problem.

***Table 5.*** *Results on the SOTAMD benchmark.*

| Test | Algorithm | EER | BPCER (%) at | |
|---|---|---|---|---|
| | | | APCER = 10% | APCER=5% |
| P&S | *Proposed approach* | 37.10% | 100.00% | 100.00% |
| | [22] | 48.04% | 85.86% | 97.35% |
| | [25] | 54.37% | 94.89% | 98.27% |
| | [31] | 43.34% | 100.00% | 100.00% |
| Digital | *Proposed approach* | 38.99% | 100.00% | 100.00% |
| | [22] | 44.81% | 100.00% | 100.00% |
| | [25] | 31.80% | 65.00% | 79.33% |
| | [31] | 41.38% | 100.00% | 100.00% |

## 7. Conclusions

In this work, different network architectures have been used for single image face morphing detection in both digital and P&S scenarios. In particular, we focused on P&S images, which still represents a big challenge today. Our initial experiments on the $MorphDB_D$ dataset proved that good performance can be achieved on digital images (BPCER=0% at APCER=10%), confirming the effectiveness of different networks already discussed in [17] [25] [26]. Unfortunately such low error rates cannot be extended to P&S images (BPCER about 12% at APCER=10% on $MorphDB_{P\&S}$) if only digital images are used for training. To overcome this problem, an automatic generation procedure has been proposed to simulate the typical P&S image degradation. When combined with automatic morphing generation it allows to produce a vast amount of training data for network training/tuning without the costs/efforts needed to manually print and scan face images. The use of simulated P&S images allowed to significantly improve morphing detection performance, achieving a BPCER=2.3% at APCER=10% on the $MorphDB_{P\&S}$ dataset. As to the different network architectures analysed, the limited size of our training databases does not allow to train large models from scratch, so all the CNN used in this work were pre-trained. The experiments highlighted that CNN pre-trained on natural images (ImageNet) can perform well on digital images, while CNN specifically pre-trained on face images (VGG Face datasets) perform better on P&S images. We argue that to detect textural differences between bona fide and morphed (digital) images, the filters learned from natural images are quite good, while in presence of P&S images more sophisticated and face-specific filters are necessary to detect the fine artifacts that survive the printing and scanning process.

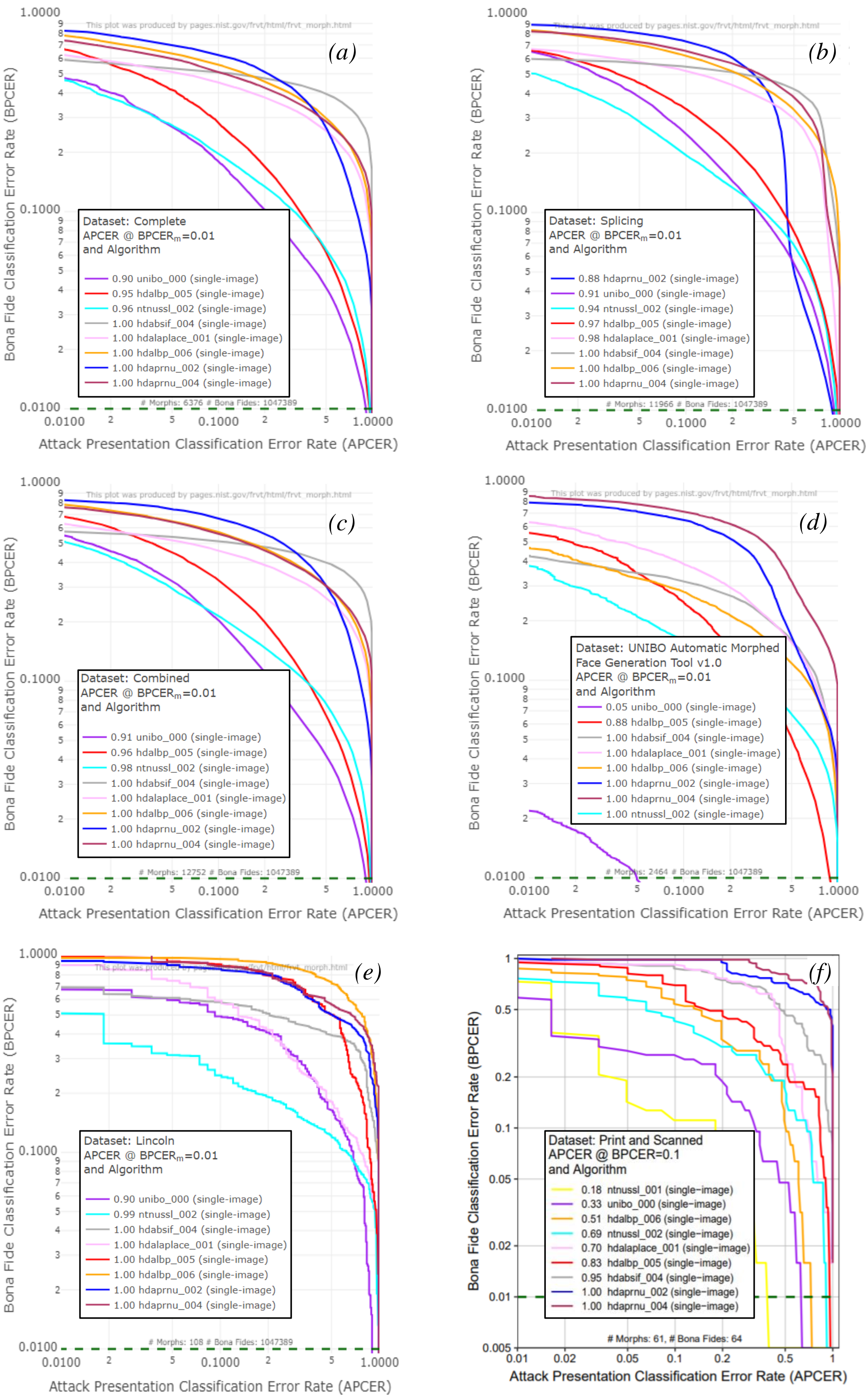

***Figure 11.*** *DET plots reporting BPCER as a function of APCER for different testing subsets. The horizontal dotted dark green line represents BPCER=0.01. The proposed algorithm corresponds to the violet curve (unibo_000).*

The tests on the NIST and SOTAMD benchmarks confirm for several data subsets the superiority of the proposed approach over other existing solutions, but generally the results obtained are quite modest. The complexity of those two benchmarks is high and single-image based morphing detection approaches struggle to reach decent performance. Morphing detection from single images has therefore to be considered a still open challenge and the unsatisfactory results suggests the importance of a very robust training, which can only be realized increasing the variability and representativeness of training data. For this reason a direct extension of our work will be to further increase the training set, using for instance different morphing algorithms or different sets of parameters for the P&S process.

Another important point to investigate in our future works is to gain some insight about the factors influencing the network decision. Some preliminary works [27] analysed the importance of different face regions for morphing detection on digital images. Further studies are necessary to better understand and explain these phenomena especially on P&S images: we believe that existing visualization techniques (see [59]) can be profitably used to this purpose.

Finally, recently Generative Adversarial Networks (GAN) [60] have been successfully used for various image generation applications (e.g., [61] [37]); their adoption for morphed image generation and P&S simulation will be investigated in our future researches.